\RequirePackage{fix-cm}
\RequirePackage{amsmath}
\documentclass[
]{svjour3}                     
\usepackage{luatex85}
\smartqed  
\usepackage{graphicx}
\usepackage[T1]{fontenc}
\usepackage{hyperref}

\usepackage{tabularx}
\newcolumntype{C}{>{\centering\arraybackslash}X}

\usepackage{multirow}
\usepackage{amssymb}
\usepackage{textcomp}
\usepackage{subfig}
\usepackage{tikz}
\usepackage[utf8]{inputenc}

\usepackage{xcolor}
\definecolor{darkgreen}{RGB}{77,175,74}
\definecolor{darkyellow}{RGB}{255,182,51}
\definecolor{myblue}{RGB}{55,126,184}

\journalname{IJCARS}
\begin{document}

\title{A Dataset of Laryngeal Endoscopic Images with Comparative Study on Convolution Neural Network Based Semantic Segmentation}

\titlerunning{Laryngeal Dataset for Comparative Study on CNN Based Semantic Segmentation}        

\author{Max-Heinrich Laves \and
        Jens Bicker \and
        Lüder A. Kahrs \and
        Tobias Ortmaier
}


\institute{Max-Heinrich Laves \and Jens Bicker \and Lüder A. Kahrs \and Tobias Ortmaier \at
           Appelstra\ss{}e 11A \\
           30167 Hannover, Germany \\
           Tel.: +49\,511 762\,19617 \\
           Fax: +49\,511 762\,19976 \\
           \email{laves@imes.uni-hannover.de}           
}

\date{Received: date / Accepted: date}

\maketitle

\begin{abstract}
  \textit{Purpose} Automated segmentation of anatomical structures in medical image analysis is a prerequisite for autonomous diagnosis as well as various computer and robot aided interventions.
Recent methods based on deep convolutional neural networks (CNN) have outperformed former heuristic methods.
However, those methods were primarily evaluated on rigid, real-world environments.
In this study, existing segmentation methods were evaluated for their use on a new dataset of transoral endoscopic exploration.\\
\textit{Methods} Four machine learning based methods SegNet, UNet, ENet and ErfNet were trained with supervision on a novel 7-class dataset of the human larynx.
The dataset contains 536 manually segmented images from two patients during laser incisions.
The Intersection-over-Union (IoU) evaluation metric was used to measure the accuracy of each method.
Data augmentation and network ensembling were employed to increase segmentation accuracy.
Stochastic inference was used to show uncertainties of the individual models.
Patient-to-patient transfer was investigated using patient-specific fine-tuning.
\\
\textit{Results} In this study, a weighted average ensemble network of UNet and ErfNet was best suited for the segmentation of laryngeal soft tissue with a mean IoU of 84.7\,\%. The highest efficiency was achieved by ENet with a mean inference time of 9.22\,ms per image.
It is shown that 10 additional images from a new patient are sufficient for patient-specific fine-tuning.
\\
\textit{Conclusion} CNN-based methods for semantic segmentation are applicable to endoscopic images of laryngeal soft tissue.
The segmentation can be used for active constraints or to monitor morphological changes and autonomously detect pathologies.
Further improvements could be achieved by using a larger dataset or training the models in a self-supervised manner on additional unlabeled data.
\keywords{Computer Vision \and Larynx \and Vocal folds \and Soft tissue \and Open access dataset \and Machine learning \and Patient-to-patient fine-tuning}
\end{abstract}

\section{Introduction}
\label{sec:introduction}

It is anticipated that the examination of laryngeal endoscopic images will allow early detection of pathologies \cite{barkmeier-kraemer2016}.
Vocal folds, as the main functional organ within the larynx, are sensitive structures for surgery. Computer vision has the potential to assist the physician in restoring or preserving the voice. This can be achieved by combining augmented reality \cite{schoob2016a}, robotics \cite{friedrich2015} and laser surgery \cite{schoob2017} with image processing methods.
Segmentation of laryngeal images is one of the most important components for successful application of such a system.

\subsection{Outline}
\label{sec:outline}

This paper discusses the automated segmentation of laryngeal soft tissue.
Therefore, we investigate state-of-the-art segmentation methods, all based on deep convolutional neural networks, and compare them on a novel manually annotated in vivo dataset of human vocal folds.
Subsequent to the description of the implementation differences of the selected models, the test setup is defined and results are reported.
The article concludes with a discussion of limitations of the chosen approaches and gives an outlook on possible improvements.

\subsection{Related Work}
\label{sec:relatedwork}

Image segmentation is the task of partitioning an image into several non-in\-ter\-sec\-ting \emph{coherent} parts and is an important step of early vision~\cite{pal1993}.
The ultimate goal in medical image segmentation is recognizing real-world objects in image data fully automatically, with high accuracy and efficiency.
However, early methods were not capable of doing this and demanded a human supervisor to initialize a method, check the result, or even to correct the segmentation afterwards \cite{olabarriaga2001}.
These techniques are based on manually selected low-level image characteristics such as grayscale thresholds, pixel color, edge detection or region growing \cite{doignon2005,osmaruiz2008,phung2005} and are therefore strongly affected by image noise or illumination changes \cite{pal1993}.
Additionally, these procedures do not perform a semantic assumption of the segmented areas.

More sophisticated methods are based on mathematical models and rely on finding the parameters with which the output of the model minimizes an error or energy function.
Some of these approaches require ground truth examples prior to or during the segmentation task.
In atlas-based segmentation, for example, a manually segmented ``atlas'', which acts as a-priori anatomical information, is matched onto the input image \cite{cabezas2011}.
This turns the segmentation problem to a registration problem.
The error-minimizing transformation between the images is also applied to the a-priori segmentation, which results in the segmentation of the input image.
A good result requires a large atlas database, which must be considered during segmentation time, thus making atlas-based segmentation not applicable in environments with real-time demands.

Artificial neural networks (ANN) have been used in medical image segmentation for a long time and are characterized by their robustness to image noise and real-time capable output due to their massively parallel structure~\cite{cabezas2011,noble2006,osmaruiz2008,pal1993,rajab2004}.
The performance of early ANNs with low numbers of neural layers were not superior compared to other methods. However, due to the recent progress in the field of convolutional neural networks, new segmentation methods with outstanding performance have been created. On image classification tasks, CNN-based approaches already exceed human-level performance~\cite{he2015}. Long et al. first proposed a fully convolutional network (FCN) which was trained pixel-to-pixel and exceeded the former state-of-the-art by up to 20\,\% on the PASCAL VOC 2012 dataset~\cite{long2015}. This led to the emergence of many new CNN-based segmentation methods. In the following, we will focus on potential real-time capable network architectures, as the segmentation result will be used later for intra-operative robot and laser control. Therefore, we will subsequently discuss four selected methods, which already showed promising results on other datasets.

One focus of this research includes the automatic detection of pathological and non-pathological areas.
In laryngeal scenes, segmentation on high-speed or stroboscopic videos is done to automatically extract the contours of the vocal folds to analyze the fold vibrations \cite{allin2004} and the glottal space to characterize glottal closure and other vocal fold pathologies~\cite{osmaruiz2008}.
Researchers have developed algorithms for the classification of laryngeal tumors based on narrow band imaging (NBI) and surrounding blood vessel structures~\cite{barbalata2016}.
Others have used classifications based on a combination of the vocal fold shape and vascular pattern \cite{turkmen2015}.
It has also been attempted to correlate voice pathologies with endoscopic videos of the vocal fold~\cite{panek2015}.
Furthermore, high-speed videos are analyzed with wavelet-based phonovibrograms and it is shown that a distinction between malignant and precancerous vocal fold lesions is possible~\cite{unger2015}.
Recently, automatic detection of vocal fold tumor tissue has been performed with confocal laser endomicroscopy and convolutional neural networks \cite{aubreville2017}.
Nevertheless, we see a lack of publicly available datasets of laryngeal (vocal fold) micro- and endoscopic images associated with comparative segmentation results.

\section{Materials and Methods}
\label{sec:methods}

First, our dataset of segmented vocal fold images will be presented.
Second, the machine learning methods used in this study are briefly described and compared.
At the end of this section, the evaluation environments are defined.

\subsection{The Vocal Folds Dataset}
\label{sec:dataset}

\begin{figure*}
    \centering
    \def\arraystretch{0.9}
    \def\resultssinglewidth{1.88cm}
    \setlength{\tabcolsep}{0.15em}
    \begin{tabularx}{\textwidth}{Xcccccc}
        \raisebox{0.075\textwidth}{\rotatebox[origin=c]{90}{images\vphantom{g}}} &
        \includegraphics[width=\resultssinglewidth]{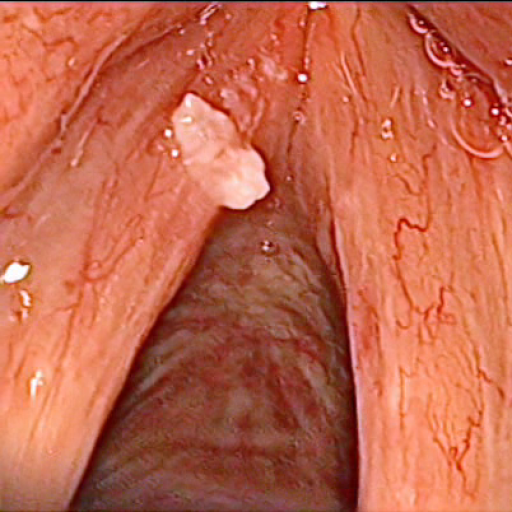} &
        \includegraphics[width=\resultssinglewidth]{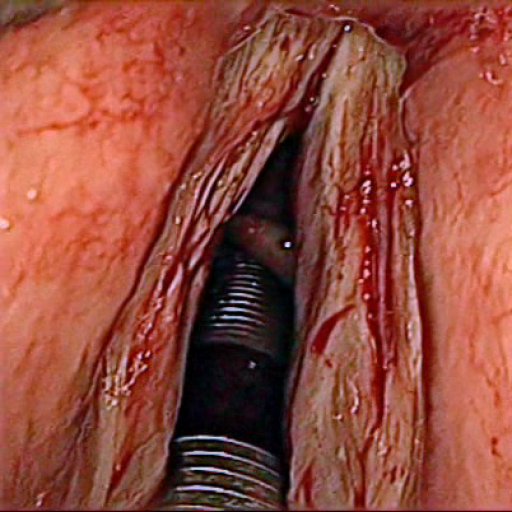} &
        \includegraphics[width=\resultssinglewidth]{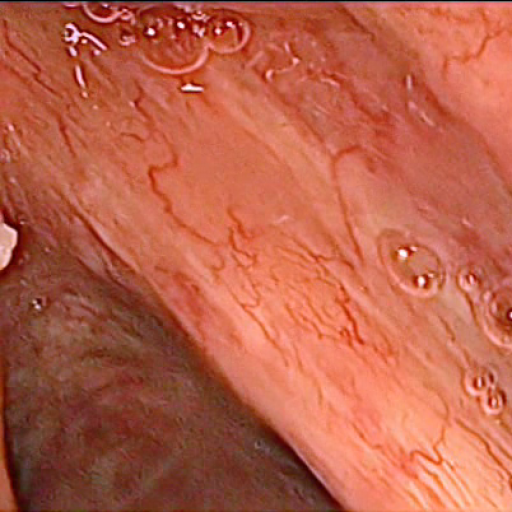} &
        \includegraphics[width=\resultssinglewidth]{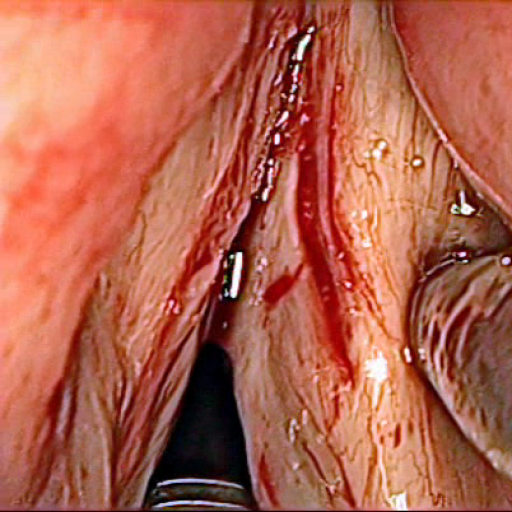} &
        \includegraphics[width=\resultssinglewidth]{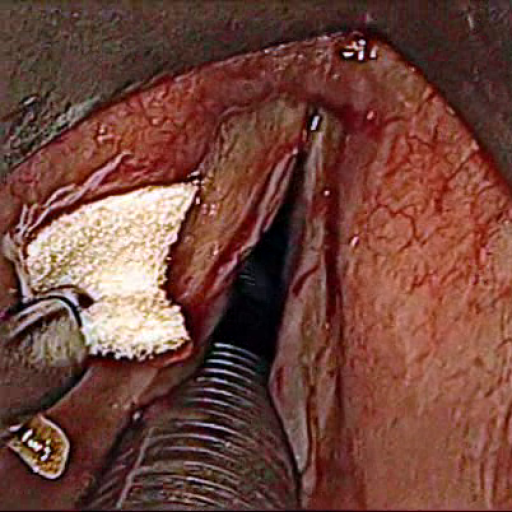} &
        \includegraphics[width=\resultssinglewidth]{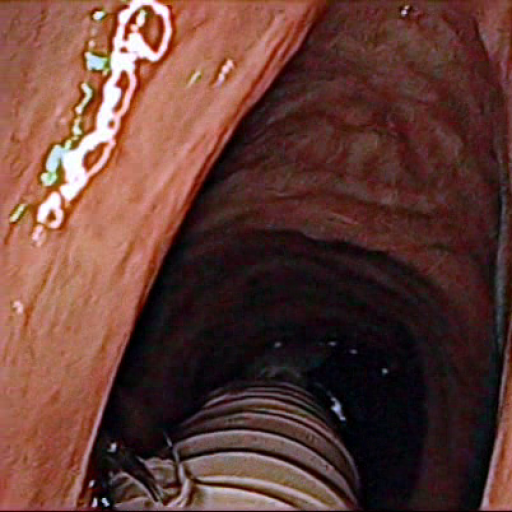} \\
        \raisebox{0.075\textwidth}{\rotatebox[origin=c]{90}{labels\vphantom{g}}} &
        \includegraphics[width=\resultssinglewidth]{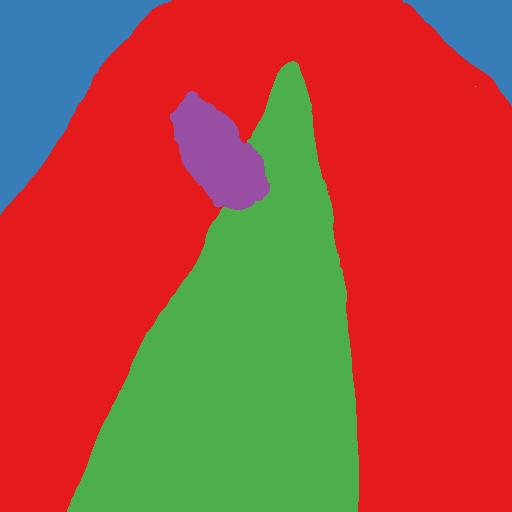} &
        \includegraphics[width=\resultssinglewidth]{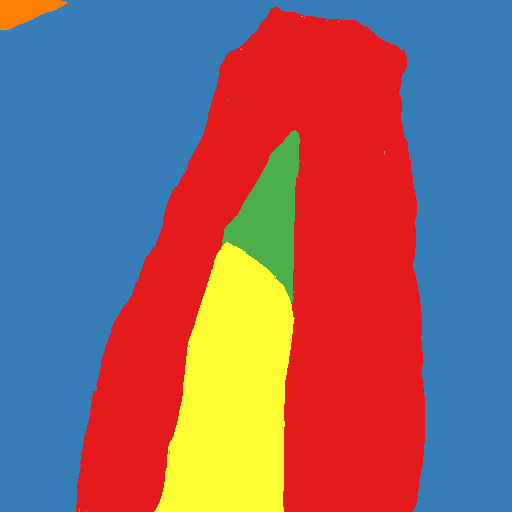} &
        \includegraphics[width=\resultssinglewidth]{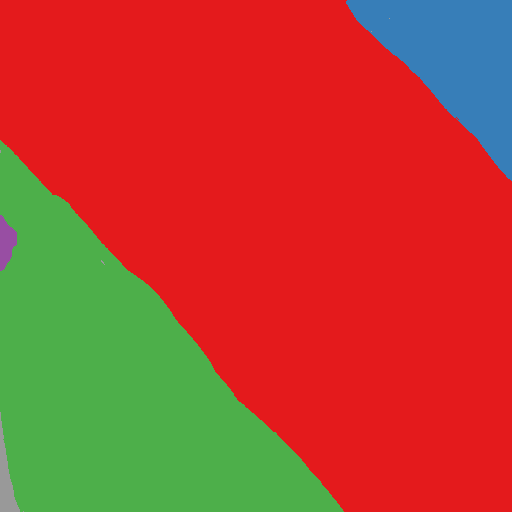} &
        \includegraphics[width=\resultssinglewidth]{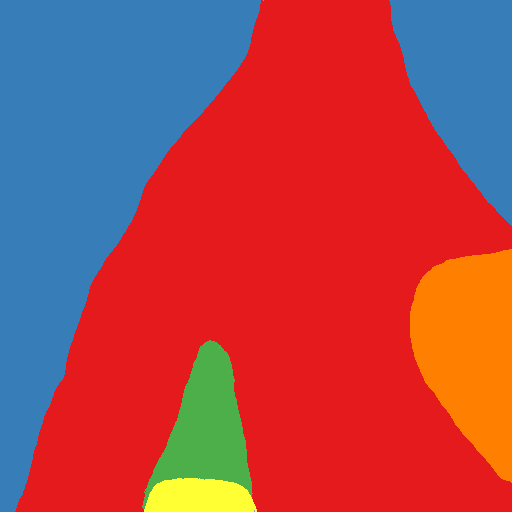} &
        \includegraphics[width=\resultssinglewidth]{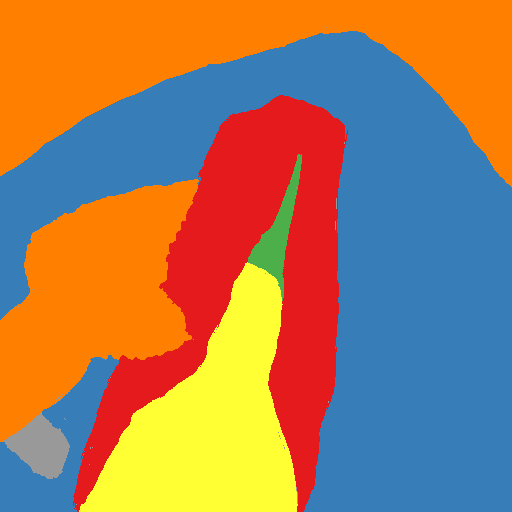} &
        \includegraphics[width=\resultssinglewidth]{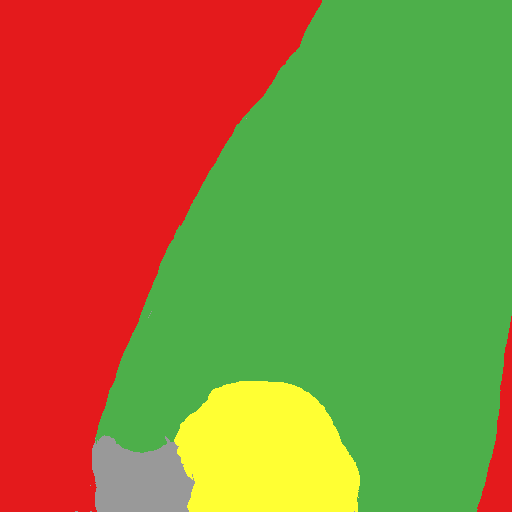}
    \end{tabularx}
    \caption{First row: Examples from vocal folds dataset. Second row: Manually segmented ground truth label maps with classes \emph{vocal folds} (red), \emph{other tissue} (blue), \emph{glottal space} (green), \emph{pathology} (purple), \emph{surgical tool} (orange), \emph{intubation} (yellow) and \emph{void} (gray). The grayscale label maps have been colorized for better visualization.}
    \label{fig:dataset}
\end{figure*}

\begin{figure}
    \centering
    \includegraphics[scale=1.0]{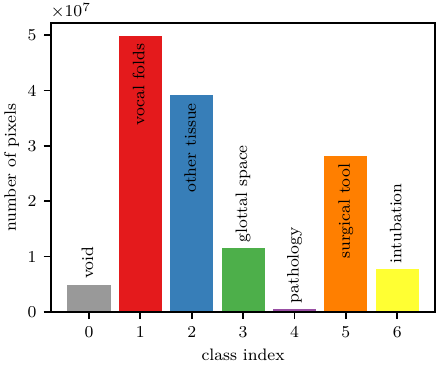}
    \caption{Number of annotated pixels per class in the dataset.}
    \label{fig:pixel_count}
\end{figure}

All subsequently described models are trained and evaluated on a dataset, containing 536 manually segmented in vivo color images of the larynx during two different resection surgeries with a resolution of $ 512 \times 512 $ pixels.
The images have been captured with a stereo endoscope (VSii, Visionsense, Petach-Tikva, Israel) in an in vivo laryngeal surgery and have been used in prior studies \cite{schoob2016}.
They are categorized in the 7 different classes \emph{void}, \emph{vocal folds}, \emph{other tissue}, \emph{glottal space}, \emph{pathology}, \emph{surgical tool} and \emph{intubation} with indices \{0, 1, 2, 3, 4, 5, 6\}, respectively, which is represented by the gray values of the label maps (see Fig~\ref{fig:dataset}).
 The dataset consists of 5 different sequences from two patients (named SEQ1--4 from patient 1 and SEQ5--8 from patient 2).
 The sequences have following characteristics:
\begin{itemize}
	\item SEQ1: pre-operative with clearly visible tumor on vocal fold, changes in translation, rotation, scale, no instruments visible, without intubation
	\item SEQ2: pre-operative with clearly visible tumor, visible instruments, changes in translation and scale, with intubation
	\item SEQ3--4: post-operative with removed tumor, damaged tissue, changes in translation and scale, with intubation
	\item SEQ5--7: pre-operative with instruments manipulating and grasping the vocal folds, changes in translation and scale, with intubation
	\item SEQ8: post-operative with blood on vocal folds, instruments and surgical dressing, with intubation
\end{itemize}
Subsequent images have a temporal contiguity as they are sampled uniformly from videos.
To reduce inter-frame correlation, images were extracted from the original videos only once per second.
In the comparative study SEQ4--SEQ6 were not used due to high similarity to SEQ3 and SEQ7 respectively, as they do not offer any additional variance to the dataset.
Segmentations have been manually created on a pen display (DTK-2241, K. K. Wacom).
Figure~\ref{fig:pixel_count} shows the distribution of the annotated pixels per class.
The dataset is publicly available\footnote{https://github.com/imesluh/vocalfolds} and will be extended in the future.

\subsection{Network Architectures}
\label{sec:models}

UNet is a fully convolutional U-shaped network architecture for biomedical image segmentation \cite{ronneberger2015}. The first part is inspired by FCN~\cite{long2015} and acts as an encoder.
It consists of repeated $ 3 \times 3 $ convolutions.
The feature maps are saved for later use before downsampling with $ 2 \times 2 $ max-pooling.
Dropout is applied at the end of the encoder.
To generate a dense segmentation output, unlike simple upsampling in FCN, the encoder is inverted, and pooling is replaced by transposed convolution (up-conv).
This acts as a decoder and creates a nearly symmetrical encoder-decoder network.
In addition, the stored encoder feature maps are concatenated after upsampling and fed into the convolution of the decoder.
UNet has a large number of feature maps and therefore the largest number of learnable parameters of the networks investigated in this study (see Tab.~\ref{tab:num_params}).
Unlike the original authors, padding of 1 and a final upsampling layer is added to the encoder to obtain matching input and output dimensions.

SegNet is a deep, fully convolutional neural network with a symmetrical encoder and decoder architecture~\cite{badrinarayanan2017}.
Each encoder layer in the network performs $ 3 \times 3 $ convolutions to create a set of feature maps.
After batch normalization, the feature maps are max-pooled to downsample.
The indices of the max-pooling layer are saved for later upsampling.
On the other side, the corresponding decoder first upsamples its input by using the saved max-pooling indices from encoding.
In contrast to UNet, transferring only max-pooling indices instead of full feature maps reduces memory consumption.
For later stochastic inference, dropout layers with $ p_{i} = 0.5 $ after every pooling/upsampling layer in encoder and decoder stages are added.
The resulting architecture is also referred to as Bayesian SegNet~\cite{kendall2017}.



ENet is an architecture with a reduced number of parameters, especially created for low latency tasks in mobile or embedded devices \cite{paszke2016}.
In contrast to SegNet, it has an asymmetric structure with a large encoder and a small decoder.
The network consists of a repeating basic module, which first splits its input into two branches.
The main branch performs three convolutions with batch normalization and a final spatial dropout with $ p_{i} = 0.01 $ in the first stage and $ p_{i} = 0.1 $ afterwards.
The outer $ 1 \times 1 $ convolutions reduce and increase the number of feature maps respectively, forming a ``bottleneck'' around the inner convolution.
The side branch just copies or copies and downsamples (max-pools) the input, which acts as a shortcut to the main branch.
As in SegNet, the max-pooling indices are saved for later upsampling by max-unpooling.
In the end, both branches are reunited by element-wise addition.
ENet has the lowest number of trainable parameters in this study (see Tab.~\ref{tab:num_params}).



ErfNet stands for ``Efficient Residual Factorized Network'' and tries to provide a compromise between accuracy and efficiency \cite{romera2018}.
It is composed of layers, which are similar to the aforementioned bottleneck modules, but instead
the 2D convolutions are factorised into 1D convolutions.
The resulting layers are called non-bottleneck-1D and drastically reduce computational cost and the number of parameters.
The network architecture
forms an encoder-decoder structure similar to ENet.
The downsampling modules are the same as the initial module of ENet.
Dropout is included in all non-bottleneck-1D layers with $ p_{i} = 0.3 $ in the last stage and $ p_{i} = 0.03 $ before of that.
Instead of max-unpooling, the upsampling block contains transposed convolution.

\subsection{Evaluation Setup}
\label{sec:testsetup}

In order to evaluate the previously described segmentation methods, performance were assessed on our in vivo vocal fold dataset.
At first, all models were implemented in Python using the PyTorch \cite{paszke2017} library with CUDA backend.
In order to train the models, the dataset was split as follows.
For reducing inter-frame correlation in the subsets, the chronologically first 50\,\% of SEQ1--3 and SEQ7--8 were used as training set (200 images), the subsequent 25\,\% as validation set (100 images) and the last 25\,\% as test set (100 images).
This seperates training and test data over time as much as possible.
The models were trained on the training set and inter-training accuracy was evaluated on the validation set.
The test set was left out for final performance assessment.
An additional training setup using only data from patient 1 is described in Section \ref{sec:pat2pat}.
As accuracy metric, the popularly accepted Intersection-over-Union~(IoU) metric is used:
\begin{equation}
    \mathrm{IoU} = \frac{\mathrm{TP}}{\mathrm{TP} + \mathrm{FP} + \mathrm{FN}}
\end{equation}
with the number of true positive (TP), false positive (FP) and false negative (FN) pixels.
The IoU metric was calculated for each of the 7 classes independently.

All models were trained in an end-to-end manner using the Adam \cite{Kingma2014} stochastic gradient descent optimizer with an initial learning rate of $ \ell = 1 \cdot 10^{-3} $, and a batch size of 6.
The learning rate was reduced by a factor of $ 10^{-1} $ when observing saturation of the validation error (reduce-on-plateau).
According to \cite{Kingma2014}, the decay rates of the first and second order moments were set to $ \beta_{1} = 0.9 $ and $ \beta_{2} = 0.999 $.
The training objective is to minimize a loss function, which acts as a dissimilarity measure between a prediction $ \hat{\vec{y}} $ of the model for input image $ \vec{x} $ and the corresponding ground truth label map $ \vec{y} $.
The prediction $ \hat{\vec{y}} $ is a tensor and expected to have the shape $ \mathbb{R}^{C \times H \times W} $ with the height $ H $, width $ W $, and number of classes $ C $ of the input image.
For every pixel of the input image, a probability distribution containing the estimated class probabilities is received.
In order to match the output dimensions of the models, one-hot encoding scheme was used, in which a ground truth label map $ \vec{y} $ is reshaped to $ \mathbb{R}^{C \times H \times W} $.
After that, the ground truth distribution for each pixel contained only one entry with 1 and otherwise 0, assigning a unique class to each pixel.
For later evaluation on the test set, the weight configuration with the lowest loss value on the validation set was chosen (early stopping) after training on a single GPU (GeForce GTX 1080 Ti, Nvidia Corp., Santa Clara, CA, USA).

As loss function, a weighted negative log-likelihood (or cross entropy) function was chosen, which is defined for a distribution $ \vec{p} = \hat{\vec{y}}_{h,w} $ of a single pixel of true class $ c $ as
\begin{equation}
    {l}(\vec{p}, c) = - \vec{w}_{c} \log\left( \frac{\exp(\vec{p}_{c})}{\sum_{j=0}^{C-1} \exp\left(\vec{p}_{j}\right)} \right)
\end{equation}
with weight vector
\begin{equation}
    \vec{w}_{c} = \frac{\sum_{j=0}^{C-1} N(j)}{C \cdot N(c)} \in \mathbb{R}^{C} ~ ,
\end{equation}
where $ N(c) $ is the total number of occurrences of pixels with class $ c $ in the whole dataset. Vector $ \vec{w} $ contains a weight for each class.
This is done to give less occurring classes, such as ``pathology'', more weight when calculating the loss.
Otherwise, omitting classes with small areas would result in a relatively small overall error.
For one prediction $ \hat{\vec{y}} $ the total loss becomes
\begin{equation}
    L( \hat{\vec{y}}, \vec{y} ) = \sum_{h=0}^{H-1} \sum_{w=0}^{W-1} l\left( \hat{\vec{y}}_{h,w}, \vec{y}_{h,w} \right) ~ .
  \end{equation}

\subsection{Transfer Learning, Data Augmentation and Network Ensembling}
\label{sec:transferaugensemble}

Additionally, all networks  were pre-trained on the publicly available Cityscapes segmentation dataset~\cite{Cordts2016}.
The Cityscapes-fine subdataset employed contains 5,000 urban steet scenes with fine annotations for 30 classes.
After pre-training, the final layer of each network was replaced by a new layer for taking the different number of classes into account and a full fine-tuning on our dataset was performed.

Data augmentation was used to increase the size of the training set to 2,000 images by horizontal flipping and rotating between -10\textdegree{} and +10\textdegree{}.
Vertical flipping and further rotations were avoided as this is unlikely in a real enviroment.
All networks were trained with and without data augmentation.

To further improve segmentation accuracy, the trained networks were combined as an anverage ensemble with trainable weights for each class and network.
Ensemble averaging was reported to significantly improve prediction accuracy and and better generalization \cite{hashem1997}.
Our combiner
\begin{equation}
  \hat{\vec{y}} \left( \vec{x} \right) = \sum_{j=1}^{p} \vec{\alpha}_{j} \circ \hat{\vec{y}}_{j} \left( \vec{x} \right)
  \label{eq:ensemble}
\end{equation}
producing the final output of the ensemble network is defined by the sum of the weighted network outputs, where $ \circ $ denotes the entrywise Hadamard product.
The weights $ \vec{\alpha}_{j} $ were optimized by fixing the weights of the individual models and training the ensemble combiner on the whole training set again.
Pairwise ensembling of the networks ($ p = 2$) and an ensemble of all networks ($ p = 4$) were investigated, which resulted in 7 different ensemble combinations.
One may think that pairwise ensembling cannot be superior to an ensemble of all networks since the combiner is trained to find the optimal combination.
However, Zhou et al. \cite{Zhou2002} have shown that ensembling \emph{many} instead of \emph{all} available models can provide better results.

\subsection{Model Uncertainty}
\label{sec:modeluncert}
In safety-critical tasks, such as medical imaging, it is important to know the confidence of the prediction of a model, especially when using the information for (semi-)autonomous robot or laser control.
According to \cite{gal2016}, the class probabilities produced by a softmax function approximate relative probabilities between class labels and give no overall measure of the model's uncertainty.
Therefore, prediction uncertainty was evaluated by Monte Carlo sampling with dropout at test time, following the framework of \cite{kendall2017} (also called \emph{stochastic inference}).
This gives an approximation of the distribution of the softmax class probabilities for every pixel of the model prediction.
The \emph{variance} $ \sigma^{2}_{c,h,w} $ of each softmax probability was used as a per-class uncertainty and the \emph{mean of the variances} $ \bar{\sigma}^{2}_{h,w} $ with respect to $ c $ as an overall uncertainty.
During Monte Carlo sampling the dropout probabilities were set to be the same as during training.

\subsection{Patient-to-Patient Generalization}
\label{sec:pat2pat}

In order to assess how well patient-independent generalization is possible, a training on the sequences of patient 1 solely and subsequent testing on patient 2 were performed.
Since \emph{pathology} does not occur in patient 2, it was neglected for this.

To additionally follow the idea of patient-specific fine-tuning \cite{Wang2018}, a minimal number of images from the beginning of the sequences of patient 2 were added to the training set.
The models were fine-tuned on this extended dataset and tested on the latter half of sequences of patient 2.
The key idea behind this is, that with a low number of pre-operatively acquired and manually segmented images of new patients, the intra-operative segmentation accuracy can be improved patient-specific.
It has been shown that even sparsely scribbled annotations can significantly improve segmentation accuracy in this case \cite{Wang2018}.
To identify the manifold of additional images needed for patient-to-patient translation, the number of images added to the training set is increased by steps of 5.

\section{Results}
\label{sec:results}
In the following, the results of the different segmentation methods from the aforementioned evaluation setup are presented.
Table~\ref{tab:ioutest} summarize the results by showing the per-class and mean segmentation accuracy on the test set measured with the Intersection-over-Union metric.

When trained from scratch (without data augmentation and pre-training), the highest mean IoU value was achieved by UNet, which also has the highest per-class IoU for \emph{pathology} and \emph{intubation} (see Tab.~\ref{tab:ioutest} (a)).
After full fine-tuning, UNet and ErfNet achieved better results, with ErfNet now performing best (see Tab.~\ref{tab:ioutest} (b)).
However, SegNet and ENet dit not benefit from pre-training, which resulted in lower per-class mean results, except for \emph{glottal space}.
All networks showed improved results by dataset augmentation (see Tab.~\ref{tab:ioutest} (c)).
In this case, ErfNet again obtained best mean IoU.
SegNet provided the worst mean results on all test scenarios.

Although it has been shown that in medical image analysis, fine-tuning of a pre-trained CNN can outperform training from scratch \cite{Tajbakhsh2016}, varying findings were observed.
ErfNet and UNet benefited from pre-training by increaded mean IoU by approx. 5.1\,\% and 2.5\,\%, respectively.
However, this was not the case for ENet and SegNet, where a decrease in mean IoU was observed by approx. 5.3\,\% and 13.2\,\%, respectively.

In general, the class \emph{pathology} appears to be worst recognized.
This can be explained by the fact that this class has a significantly lower occurrence in the whole dataset (see Fig.~\ref{fig:pixel_count}), even lower than \emph{void}, which has not been considered for accuracy assessment.
However, good results were achieved for classes \emph{vocal folds}, \emph{other tissue}, \emph{surgical tool} and \emph{intubation}.

Table~\ref{tab:ioutest} (d) shows, that network-and-class-wise average ensembling clearly improved segmentation results.
The networks were ensembled after individual training on the augmented dataset.
Except for ENet+SegNet configuration, all ensembles performed better in comparison to their individual models.

\begin{figure*}
    \centering
    \def\arraystretch{0.9}
    \def\resultssinglewidth{1.592cm}
    \setlength{\tabcolsep}{0.15em}
    \begin{tabularx}{\textwidth}{Xccccccc}
     & \raisebox{0.3em}{input} & \raisebox{0.3em}{ground truth} & \raisebox{0.3em}{\textbf{UNet}} & \raisebox{0.3em}{\textbf{SegNet}} & \raisebox{0.3em}{\textbf{ENet}} & \raisebox{0.3em}{\textbf{ErfNet}} & \raisebox{0.3em}{\textbf{Ensemble}} \\
        \raisebox{0.06\textwidth}{\rotatebox[origin=c]{90}{SEQ2\vphantom{g}}} &
        \includegraphics[width=\resultssinglewidth]{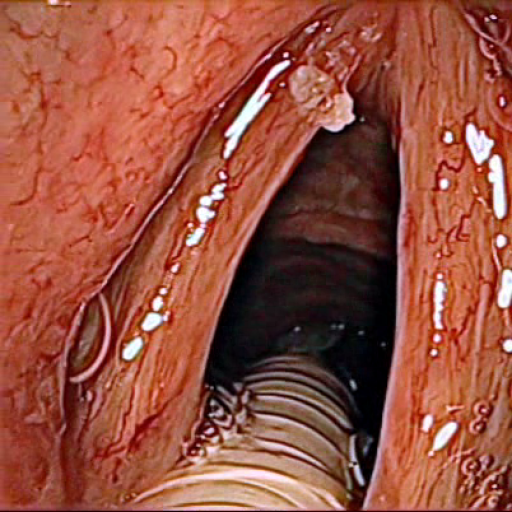} &
        \includegraphics[width=\resultssinglewidth]{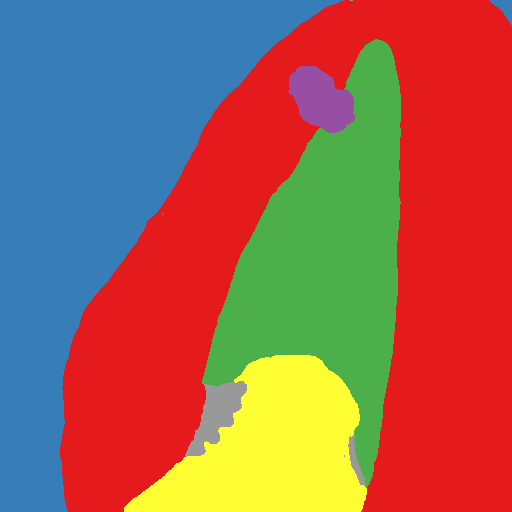} &
        \includegraphics[width=\resultssinglewidth]{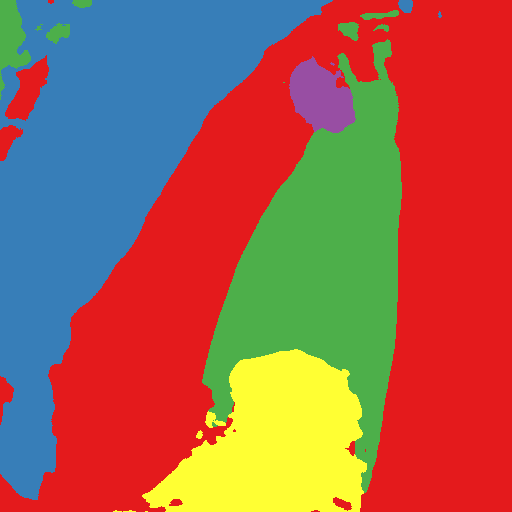} &
        \includegraphics[width=\resultssinglewidth]{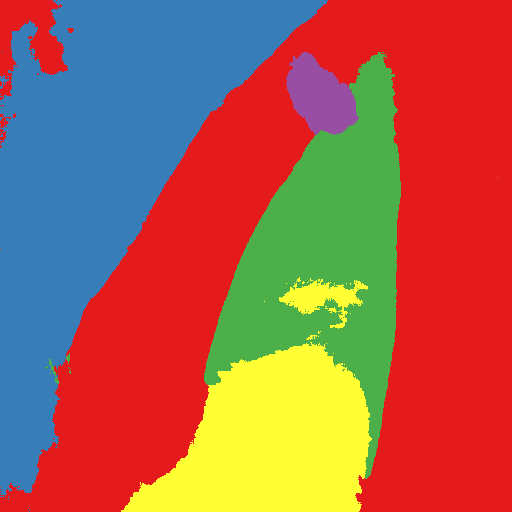} &
        \includegraphics[width=\resultssinglewidth]{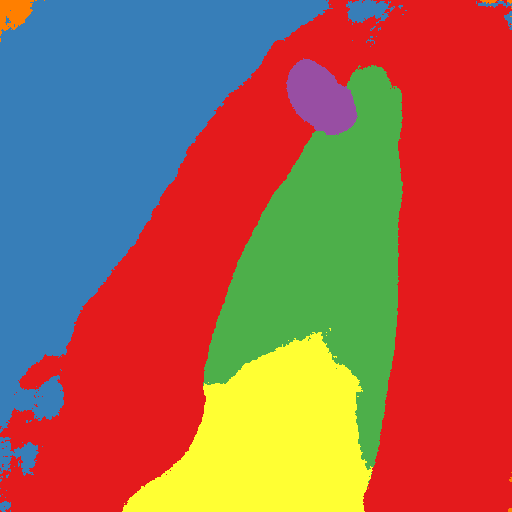} &
        \includegraphics[width=\resultssinglewidth]{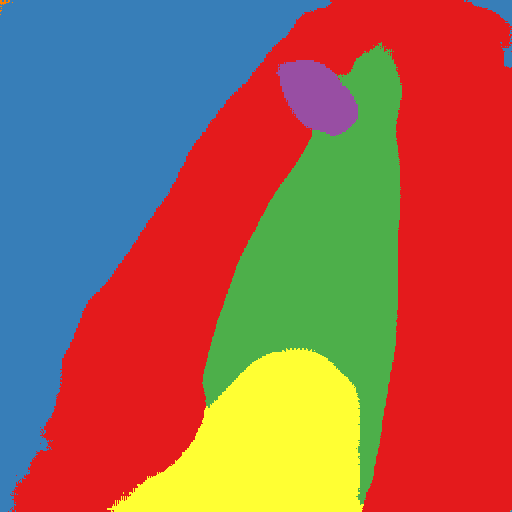} &
        \includegraphics[width=\resultssinglewidth]{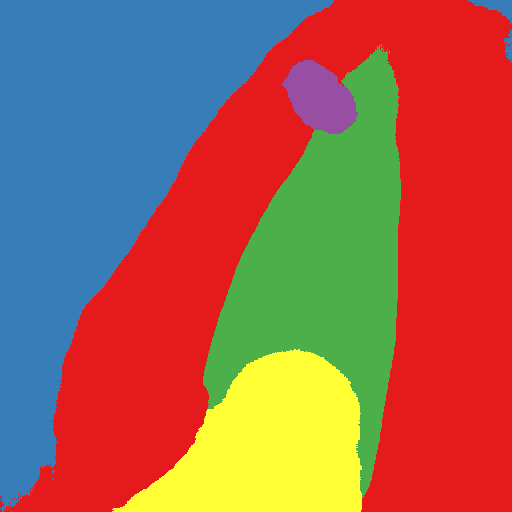} \\

        \raisebox{0.06\textwidth}{\rotatebox[origin=c]{90}{uncertainties\vphantom{g}}} & & &
        \includegraphics[width=\resultssinglewidth]{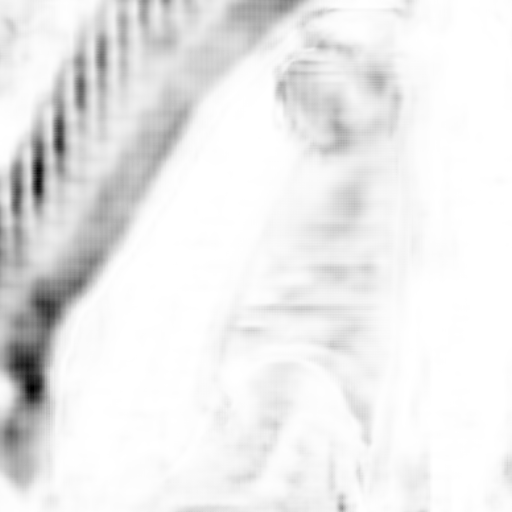} &
        \includegraphics[width=\resultssinglewidth]{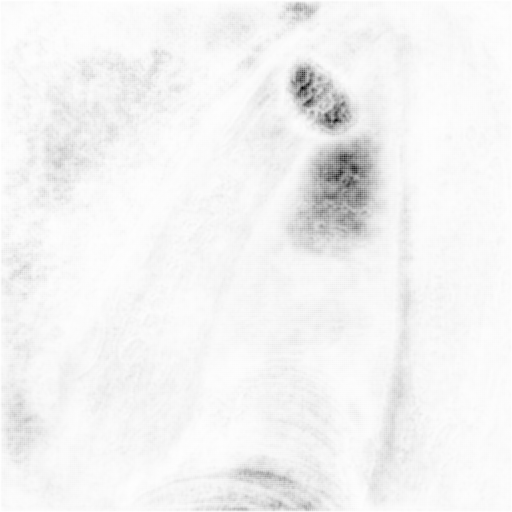} &
        \includegraphics[width=\resultssinglewidth]{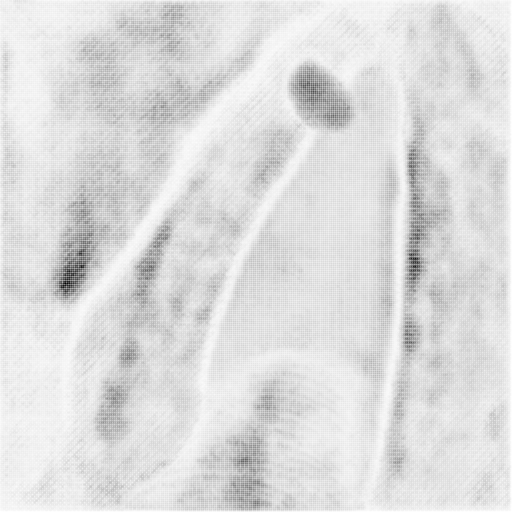} &
        \includegraphics[width=\resultssinglewidth]{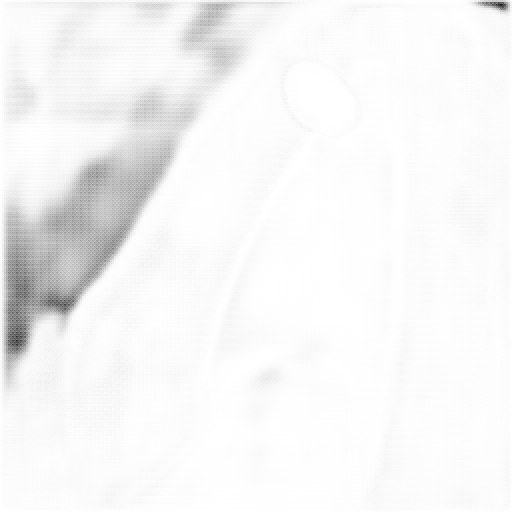} &
        \includegraphics[width=\resultssinglewidth]{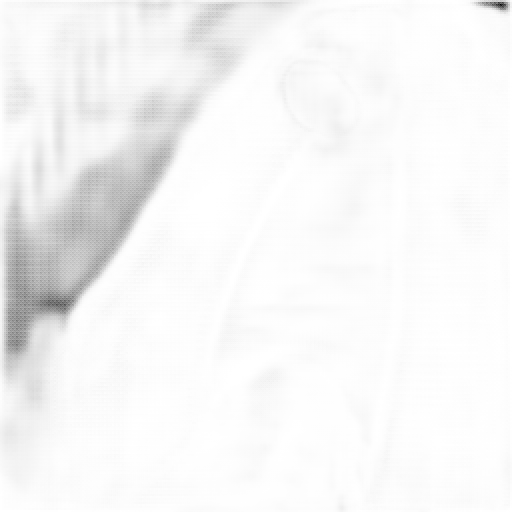} \\

        \raisebox{0.06\textwidth}{\rotatebox[origin=c]{90}{SEQ3\vphantom{g}}} &
        \includegraphics[width=\resultssinglewidth]{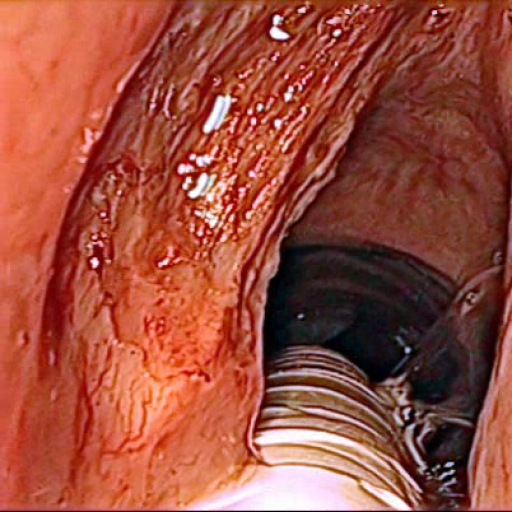} &
        \includegraphics[width=\resultssinglewidth]{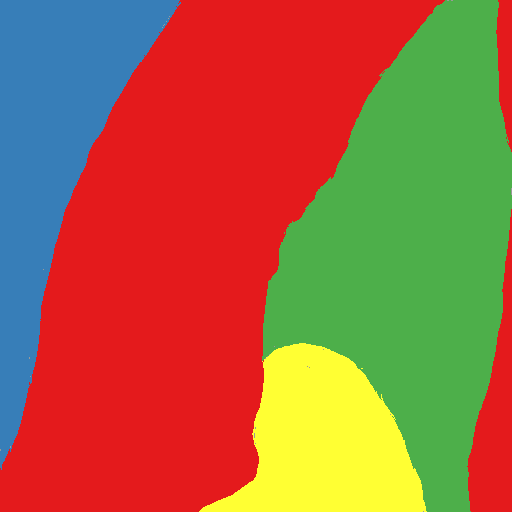} &
        \includegraphics[width=\resultssinglewidth]{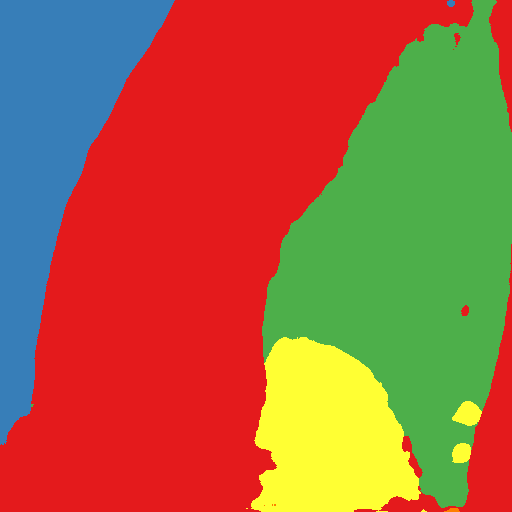} &
        \includegraphics[width=\resultssinglewidth]{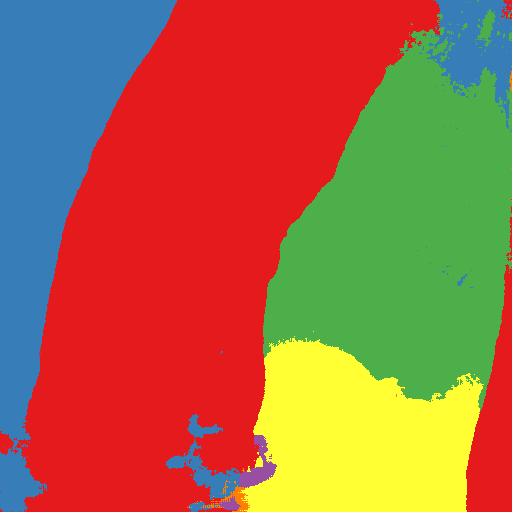} &
        \includegraphics[width=\resultssinglewidth]{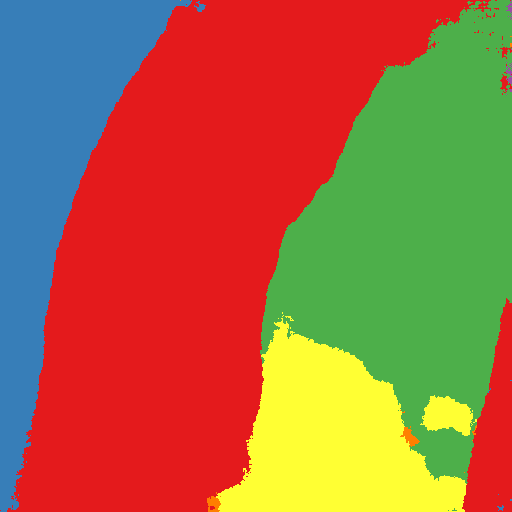} &
        \includegraphics[width=\resultssinglewidth]{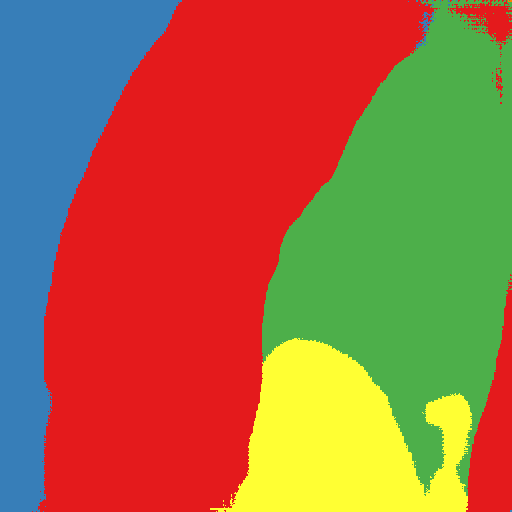} &
        \includegraphics[width=\resultssinglewidth]{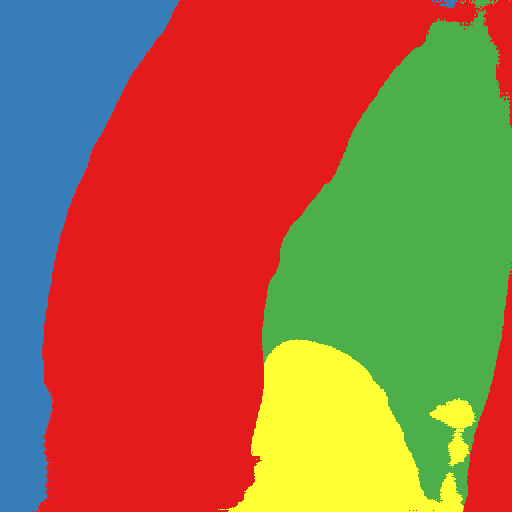} \\

        \raisebox{0.06\textwidth}{\rotatebox[origin=c]{90}{uncertainties\vphantom{g}}} & & &
        \includegraphics[width=\resultssinglewidth]{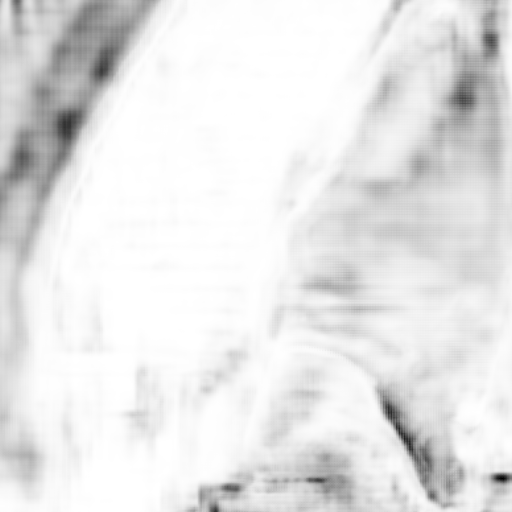} &
        \includegraphics[width=\resultssinglewidth]{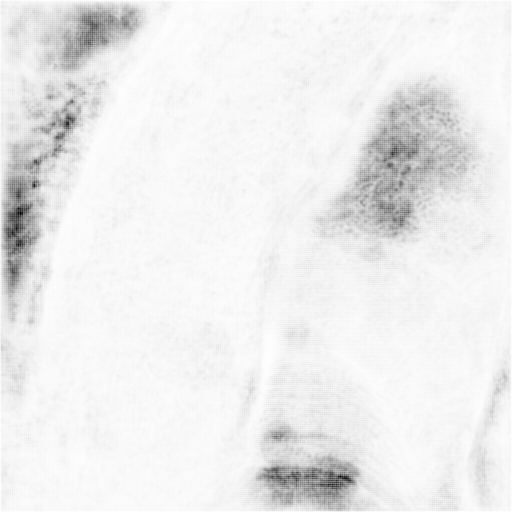} &
        \includegraphics[width=\resultssinglewidth]{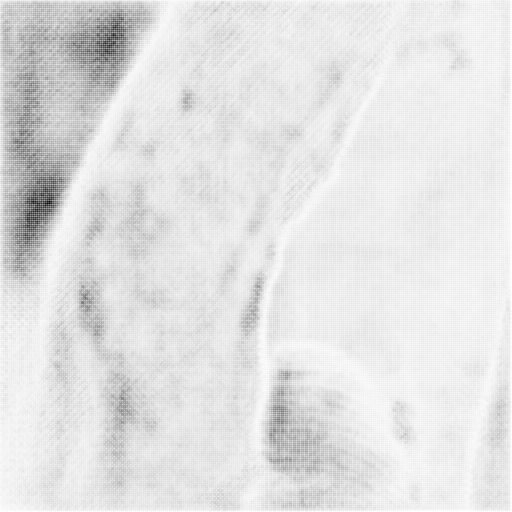} &
        \includegraphics[width=\resultssinglewidth]{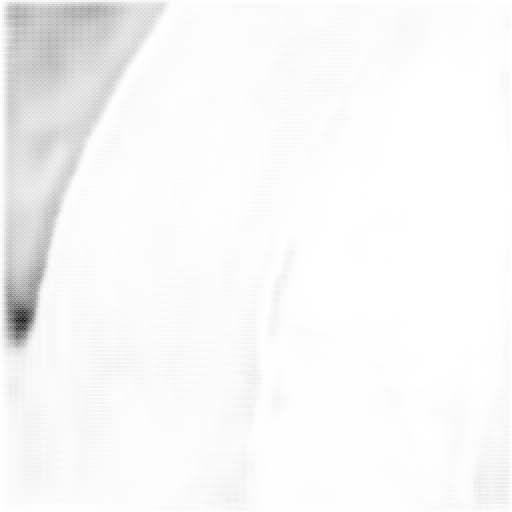} &
        \includegraphics[width=\resultssinglewidth]{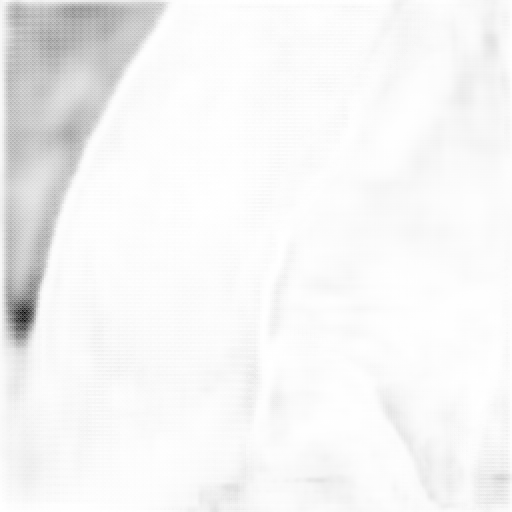} \\

        \raisebox{0.06\textwidth}{\rotatebox[origin=c]{90}{SEQ8\vphantom{g}}} &
        \includegraphics[width=\resultssinglewidth]{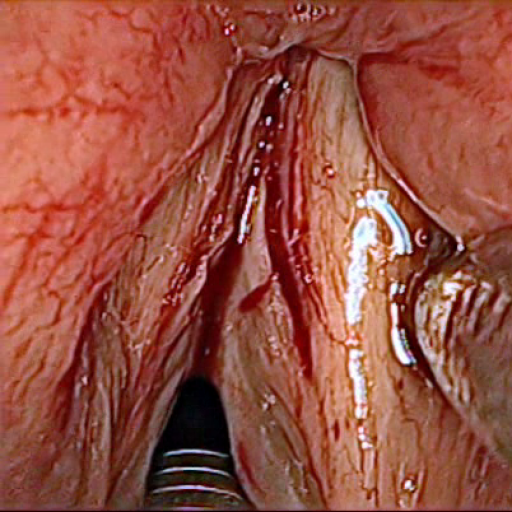} &
        \includegraphics[width=\resultssinglewidth]{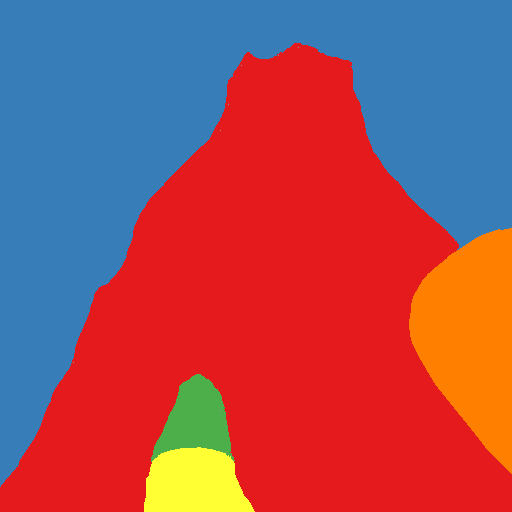} &
        \includegraphics[width=\resultssinglewidth]{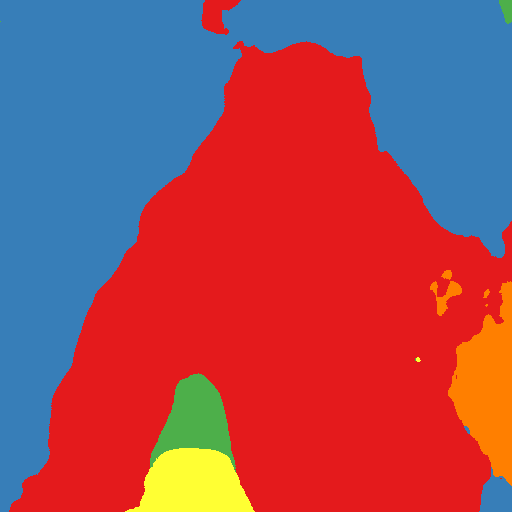} &
        \includegraphics[width=\resultssinglewidth]{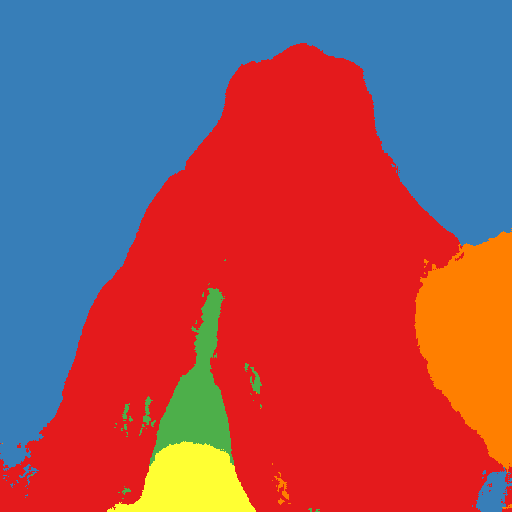} &
        \includegraphics[width=\resultssinglewidth]{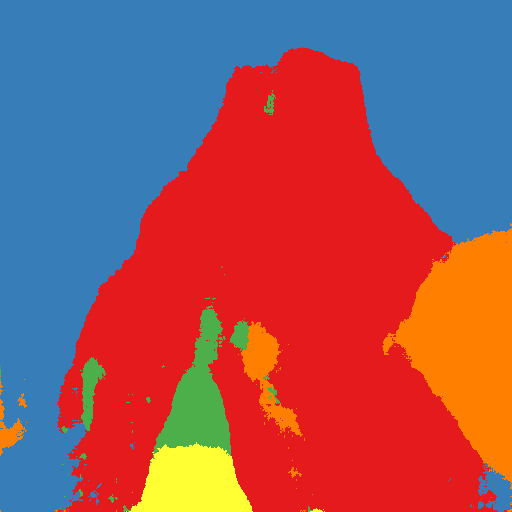} &
        \includegraphics[width=\resultssinglewidth]{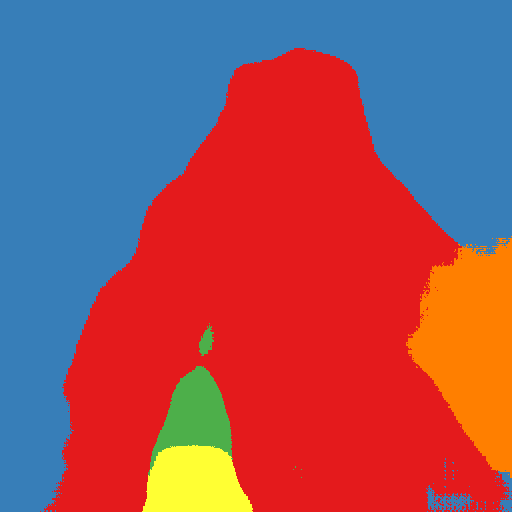} &
        \includegraphics[width=\resultssinglewidth]{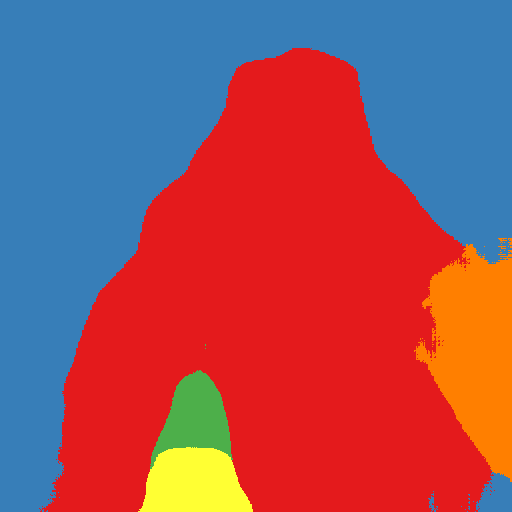}\\

        \raisebox{0.06\textwidth}{\rotatebox[origin=c]{90}{uncertainties\vphantom{g}}} & & &
        \includegraphics[width=\resultssinglewidth]{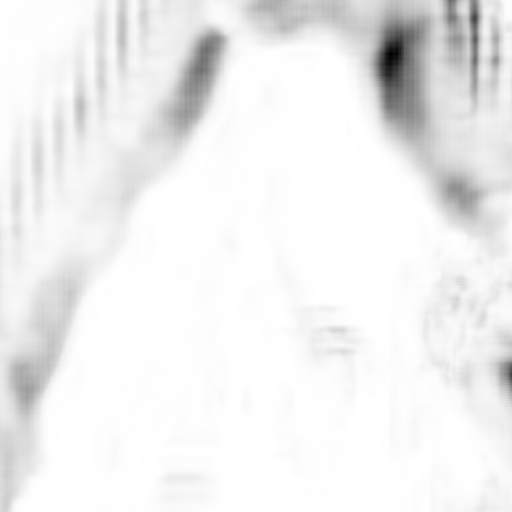} &
        \includegraphics[width=\resultssinglewidth]{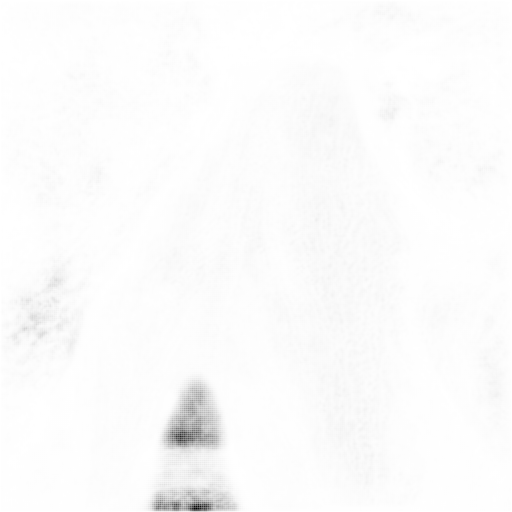} &
        \includegraphics[width=\resultssinglewidth]{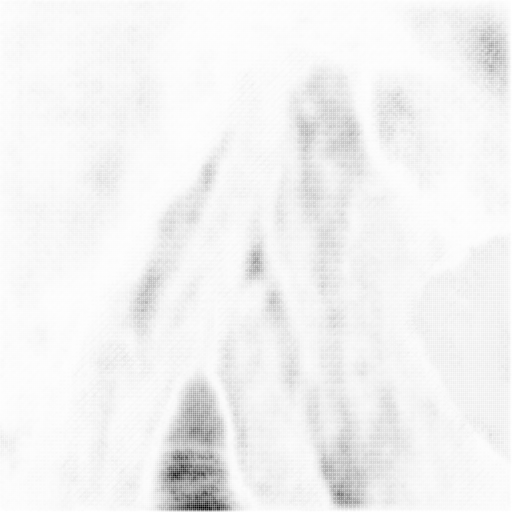} &
        \includegraphics[width=\resultssinglewidth]{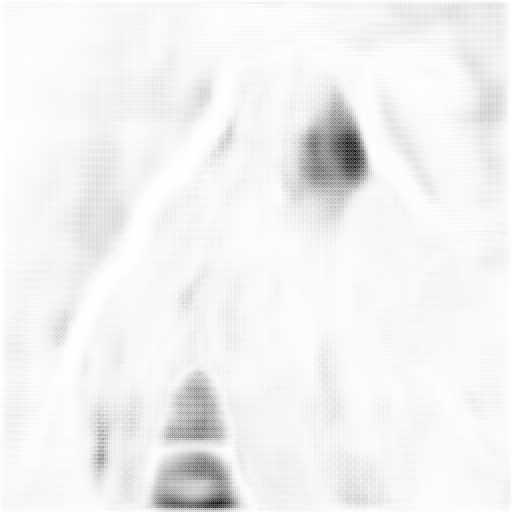} &
        \includegraphics[width=\resultssinglewidth]{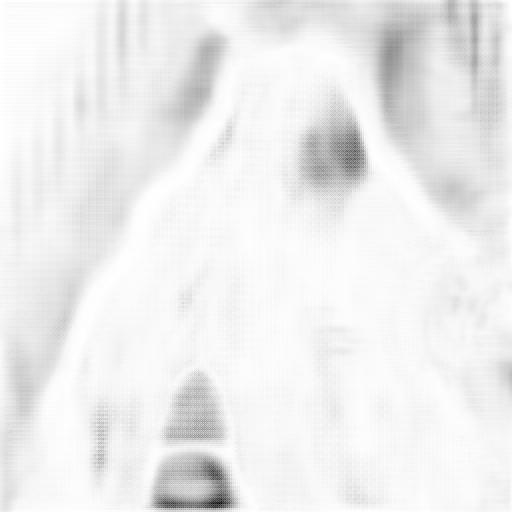}
    \end{tabularx}
    \caption{Qualitative results of the different networks on exemplary images of the test set sequences and corresponding prediction uncertainties of all classes (gray level denotes variance with white where $ \bar{\sigma}^{2}_{h,w} = 0 $). The results were generated by selecting the training state that provided best results on the validation set.}
    \label{fig:results}
\end{figure*}

Figure~\ref{fig:results} visualizes class label predictions for selected example images from the test set after training on the augmented dataset.
As expected from the IoU values, the prediction of SegNet has major errors in classes \emph{pathology}, \emph{vocal folds} and \emph{other tissue}.
The results of UNet are better in general, but it provided visible errors on the edges of two adjacent areas.
ENet and ErfNet both achieved good results, with ErfNet also resolving edges and smaller areas well.
When zooming into the label maps, it is visible that ENet has a strong aliasing effect on the edges, whereas ErfNet results in smoother edges.
The ensemble results are reported for ErfNet+UNet configuration, which provided overall best results.
Please see the supplemental video material associated with this publication.

The prediction uncertainties in Fig.~\ref{fig:results} give an estimate on how confident the model is for a specific pixel of the selected images.
UNet and SegNet use normal dropout for stochastic inference, whereas ENet and ErfNet use spatial dropout (turning off full feature maps).
In contrast to normal dropout, there is currently no proof of spatial dropout acting as Bayesian approximation \cite{gal2016}.
This must be taken into account when considering the uncertainty maps.
It is worth mentioning that ENet and ErfNet have a higher uncertainty in general and especially on \emph{other tissue}.
A possible explanation for this could be the low numbers of parameters having less redundancies compared to UNet and SegNet (see Tab.~\ref{tab:num_params}).
Furthermore, it can be observed that with the exception of ErfNet, all models show a high uncertainty for \emph{pathology}.
On the other hand, all models show low uncertainties for \emph{vocal folds}, which can be interpreted as good generalization for this class.

\begin{figure}
    \centering
    \begin{tikzpicture}
    	\draw (0, 0) node[inner sep=0] {\includegraphics[scale=1.0]{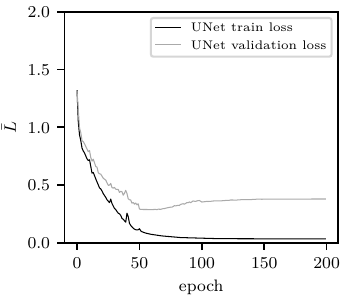}~\includegraphics[scale=1.0]{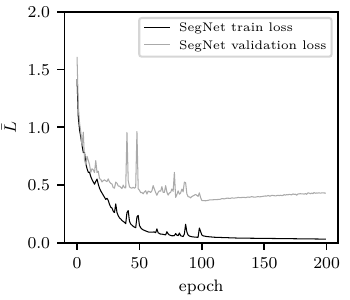}};
    	\draw (-3.441, -1.151) node {$\uparrow$};
    	\draw (3.5315, -1.0) node {$\uparrow$};
	\end{tikzpicture}
	\begin{tikzpicture}
    	\draw (0, 0) node[inner sep=0] {\includegraphics[scale=1.0]{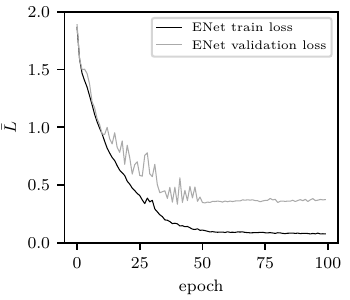}~\includegraphics[scale=1.0]{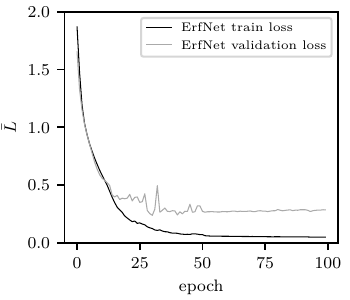}};
    	\draw (-2.963, -1.0635) node {$\uparrow$};
    	\draw (2.632, -1.248) node {$\uparrow$};
	\end{tikzpicture}
    \caption{Mean losses during training of individual networks on augmented dataset. The arrows denote selected epochs for early stopping.}
    \label{fig:loss_over_time}
\end{figure}

Figure~\ref{fig:loss_over_time} illustrates the recorded mean loss during training on both the training and validation sets.
All models converged in our training setup.
An increase in validation loss can be observed as soon as the training loss gets close to $ \bar{L} = 0 $, which indicates overfitting.
Early stopping was used to prevent this.
The corresponding epochs are marked with an arrow in Fig~\ref{fig:loss_over_time}.

\begin{figure}
    \centering
    \includegraphics[scale=1.0]{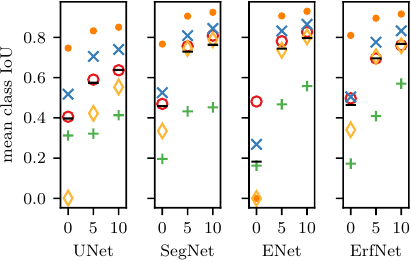}
    \caption[Results of patient-to-patient transfer and patient-specific fine-tuning.]{Results of patient-to-patient transfer and patient-specific fine-tuning. The $x$-axis indicates number of images used from patient 2. The class IoU is denoted by symbols for \emph{vocal folds} (\tikz\draw[red, thick] (0,0) circle (.5ex);), other tissue (\textcolor{myblue}{$ \mathbf{\times} $}), \emph{glottal space} (\textcolor{darkgreen}{$\mathbf{+}$}), \emph{surgical tool} (\tikz\draw[orange,fill=orange] (0,0) circle (.35ex);), \emph{intubation} (\textcolor{darkyellow}{$ \mathbf{\diamond} $}) and the mean IoU is denoted by black bar (\textbf{--}).}
    \label{fig:pat2pat}
\end{figure}
Figure~\ref{fig:pat2pat} shows the patient-to-patient generalization of the different networks and the effect of patient-specific fine-tuning.
All networks benefit greatly from additional data from patient 2, resulting in an average IoU increase of 31.0 \%-points with 5 additional images and another 5.6 \%-points with 10 additional images.
The latter results are already comparable to those in Tab.~\ref{tab:ioutest} and therefore no further images were used for patient-specific fine-tuning.

Table~\ref{tab:infertime} shows the mean inference times including memory transfer to the GPU. All models are able to process the images efficiently with at least $ 20 $ frames per second (fps).
ENet and ErfNet, however, perform significantly better, with ENet achieving the best performance ($ 108.5 $\,fps).

\begin{table*}
    \centering
\subfloat[][training from scratch]{
    \begin{tabular}{lccccccr}
        \hline
		\multirow{2}{*}{\textbf{Network}} & \multicolumn{6}{c}{\textbf{Label Index}} & \multirow{2}{*}{\textbf{mean}} \\
         & \textbf{1} & \textbf{2} & \textbf{3} & \textbf{4} & \textbf{5} & \textbf{6} & \\
        \hline
        \textbf{UNet}    & 76.3 & 74.0 & 60.0 & \textbf{64.5} & 87.3 & \textbf{79.3} & \textbf{73.6} \\
        \textbf{SegNet}  & 68.8 & 69.7 & 46.2 & 63.2 & 86.1 & 69.5 & 67.3 \\
        \textbf{ENet}    & \textbf{79.6} & \textbf{79.8} & 67.4 & 51.3 & 86.8 & 69.5 & 72.4 \\
        \textbf{ErfNet}  & 75.6 & 76.8 & \textbf{70.1} & 49.9 & \textbf{90.9} & 77.4 & 73.5 \\
        \hline
        \textbf{mean}    & 75.1 & 75.1 & 60.9 & 57.2 & 87.8 & 73.9 & \\
    \end{tabular}
    } \\
    \subfloat[][pre-training and fine-tuning]{
    \begin{tabular}{lccccccr}
        \hline
        \multirow{2}{*}{\textbf{Network}} & \multicolumn{6}{c}{\textbf{Label Index}} & \multirow{2}{*}{\textbf{mean}} \\
         & \textbf{1} & \textbf{2} & \textbf{3} & \textbf{4} & \textbf{5} & \textbf{6} & \\
        \hline
        \textbf{UNet}    & 72.6 & 71.2 & 72.4 & 71.9 & 88.8 & \textbf{79.4} & 76.1 \\
        \textbf{SegNet}  & 65.3 & 66.7 & 42.7 & 20.8 & 76.5 & 52.5 & 54.1 \\
        \textbf{ENet}    & 74.5 & 75.0 & 66.0 & 32.0 & 88.9 & 66.2 & 67.1 \\
        \textbf{ErfNet}  & \textbf{78.5} & \textbf{78.9} & \textbf{73.4} & \textbf{74.5} & \textbf{90.0} & 76.2 & \textbf{78.6} \\
        \hline
        \textbf{mean}    & 72.7 & 73.0 & 63.6 & 49.8 & 86.1 & 68.6 &  \\
    \end{tabular}
} \\
\subfloat[][training on augmented dataset]{
  \begin{tabular}{lccccccr}
        \hline
        \multirow{2}{*}{\textbf{Network}} & \multicolumn{6}{c}{\textbf{Label Index}} & \multirow{2}{*}{\textbf{mean}} \\
         & \textbf{1} & \textbf{2} & \textbf{3} & \textbf{4} & \textbf{5} & \textbf{6} & \\
        \hline
        \textbf{UNet}    & 76.9 & 75.5 & \textbf{71.9} & 64.7 & 90.1 & \textbf{81.6} & 76.8 \\
        \textbf{SegNet}  & 73.6 & 72.6 & 68.2 & 58.3 & 87.1 & 74.2 & 72.3 \\
        \textbf{ENet}    & 81.2 & 80.5 & 65.7 & 75.7 & 88.6 & 78.7 & 78.4 \\
        \textbf{ErfNet}  & \textbf{85.4} & \textbf{86.0} & 70.5 & \textbf{75.9} & \textbf{90.5} & 81.2 & \textbf{81.6} \\
        \hline
        \textbf{mean}    & 79.3 & 78.7 & 69.1 & 68.7 & 89.1 & 78.9 &  \\
    \end{tabular}
} \\
\subfloat[][ensemble configurations]{
\begin{tabular}{lccccccr}
        \hline
        \multirow{2}{*}{\textbf{Ensemble}} & \multicolumn{6}{c}{\textbf{Label Index}} & \multirow{2}{*}{\textbf{mean}} \\
         & \textbf{1} & \textbf{2} & \textbf{3} & \textbf{4} & \textbf{5} & \textbf{6} & \\
        \hline
        \textbf{ENet+ErfNet}   & \textbf{85.7} & \textbf{85.8} & 72.7 & 83.3 & 90.9 & 82.3 & 83.5 \\
        \textbf{ENet+SegNet}   & 79.2 & 78.1 & 70.3 & 69.2 & 89.2 & 80.5 & 77.8 \\
        \textbf{ENet+UNet}     & 82.3 & 81.7 & 73.8 & 84.3 & 91.7 & 84.2 & 83.0 \\
        \textbf{ErfNet+SegNet} & 83.1 & 82.9 & 73.5 & 77.9 & 90.6 & 82.6 & 81.8 \\
        \textbf{ErfNet+UNet}   & 84.9 & 84.9 & \textbf{77.1} & 84.4 & \textbf{92.5} & \textbf{84.4} & \textbf{84.7} \\
        \textbf{UNet+SegNet}   & 78.5 & 77.0 & 75.9 & 69.2 & 91.0 & 81.2 & 78.8 \\
	\textbf{All}           & 84.9 & 84.1 & 76.0 & \textbf{86.6} & 92.1 & 84.3 & \textbf{84.7} \\
        \hline
        \textbf{mean}          & 82.7 & 82.1 & 74.2 & 79.3 & 91.1 & 82.8 &  \\
    \end{tabular}
}
    \caption{Per-class and mean IoU (\%) on the test set for different training scenarios. Bold numbers denote best results.}
    \label{tab:ioutest}
\end{table*}

\begin{table}
    \centering
    \begin{tabular}{lccc}
        \hline
        \multirow{2}{*}{\textbf{Network}} & \multirow{2}{*}{trainable parameter} & \multicolumn{2}{c}{$512 \times 512$} \\
        &  & ms & fps \\
        \hline
        \textbf{UNet}    & 31,032,200 & 47.2 & 21.18 \\
        \textbf{SegNet}  & 29,447,624 & 45.7 & 21.9  \\
        \textbf{ENet}    & 392,420    & \textbf{9.22} & \textbf{108.5} \\
        \textbf{ErfNet}  & 2,064,508  & 11.1 & 90.1  \\
        \hline
    \end{tabular}
    \caption{Total number of trainable parameters and mean inference times on a single GeForce GTX 1080 Ti (including data to GPU transfer).}
    \label{tab:infertime}
    \label{tab:num_params}
\end{table}

\section{Conclusion and outlook}
\label{sec:conclusion}

In this paper, a novel dataset of laryngeal surgery images with ground truth segmentation maps has been introduced.
The dataset was used for a comparative study of recent CNN-based segmentation methods.
It has been shown that an ensemble of ErfNet and UNet yielded the most successful results on  segmenting endoscopic image data.
Schoob et al. \cite{schoob2017} reported that image-based online control of incision lasers for soft tissue undergoing motion is feasible with an imaging pipeline running at 73.5 Hz.
Therefore, both ENet and ErfNet are well-suited in terms of efficiency for later use in autonomous or robot aided interventions.
The reported laser incision error of their approach was 0.12--0.21\,mm.
With monoscopic images, we cannot directly compare our segmentation accuracy to this and will address this in future work.
However, with sufficient safety margins, segmentation can be used to define active constrains.

The raw data, from which our dataset images are chosen, consist of stereo images.
Currently, the dataset only includes the left images.
By segmenting both the left and the right stereo images, a metric accuracy value for the segmentation task can be stated, if calibration data is available.

When comparing UNet and SegNet to ENet and ErfNet, two main characteristics differ.
The first difference is that the first two have a symmetrical auto-encoder structure, while the latter two have a significantly smaller decoder.
It is assumed that a higher performance can be attributed to the encoder, while the decoder only upsamples the output of the encoder \cite{paszke2016}.
The second difference is long-term connections between encoder and decoder layers, bypassing the deeper structures, by copying whole feature maps in UNet or by memorizing the pooling indices in SegNet.
However, ENet and ErfNet do not have such connections. The residual units bypass only one layer.

With pre-training, the performance of ErfNet and UNet were improved on our dataset.
However, not all networks benefited from pre-training.
A possible explanation for this can be the different image domains, as the Cityscapes dataset contains segmented street scenes.
This underlines the need for publicly available medical segmentation datasets.
In addition, an extension of the dataset to different pathologies can enable automated diagnosis.
Further improvement regarding our dataset and the generation of ground truth information for medical imaging is required.

The disadvantage of our dataset up to now is the low number of patients and the unbalanced class occurrences.
The most important class \emph{pathology} is the least common.
Besides using a class-weighted cross entropy, specific loss functions for extreme class imbalances can be used, such as the focal loss \cite{Lin2017}.
Ideally, performance of networks in the medical imaging domain should be evaluated on data from patients not included in the training set.
Therefore, results from Tab.~\ref{tab:ioutest} can be considered for comparing segmentation performances but do not show the ability of inter-patient generalization for real clinical use.
Results from Fig.~\ref{fig:pat2pat} indicate, that training with data from one patient is not sufficient for a well-generalized segmentation model in a laryngeal environment.
However, a patient-specific fine-tuning with only 10 additional images from a new patient already seem to be sufficient for good segmentation results.
These images could be taken during a pre-operative examination and used for fine-tuning after sparse manual annotation.

Ongoing work will therefore focus on enlarging the dataset by investigating self-supervised segmentation methods, where a human expert accepts, discards or corrects the segmentation prediction.
Beyond that, improvements can be achieved by modelling the uncertainty, especially for small datasets like ours \cite{kendall2017}, by combining CNN-based segmentation with tracking \cite{garcia-peraza-herrera2017} or by using additional unlabeled data \cite{creswell2018}.

\begin{acknowledgements}
We thank Giorgio Peretti from the Ospedale Policlinico San Martino, University of Genova, Italy, for providing us with the in vivo laryngeal data used in this study.
We would also like to thank James Napier from the Institute of Lasers and Optics, University of Applied Sciences Emden-Leer, Germany, for his thorough proofreading of this manuscript.
\end{acknowledgements}

\section*{Disclosure of potential conflicts of Interest}

\paragraph{Conflict of Interest} The authors declare that they have no conflict of interest.

\paragraph{Funding} This research has received funding from the European Union as being part of the ERFE OPhonLas project.

\paragraph{Formal Consent} The endoscopic video images were acquired by Prof. Giorgio Peretti (Director of Otorhinolaryngology at Ospedale Policlinico San Martino, University of Genova). Patients gave their written consent for the procedure and the use of the data. No further approval is necessary for such endoscopic recordings. The videos were anonymized made available inside the µRALP consortium for further usage. All procedures performed in studies involving human participants were in accordance with the ethical standards of the institutional and/or national research committee and with the 1964 Helsinki declaration and its later amendments or comparable ethical standards.

\bibliographystyle{spmpsci} 
\bibliography{library}   

\end{document}